\pdfoutput=1

\PassOptionsToPackage{table}{xcolor}

\documentclass[sigconf]{acmart}

\AtBeginDocument{%
  }

\copyrightyear{2026}
\acmYear{2026}
\setcopyright{cc}
\setcctype{by}
\acmConference[MM '26]{Proceedings of the 34th ACM International Conference on Multimedia}{November 10--14, 2026}{Rio de Janeiro, Brazil}
\acmBooktitle{Proceedings of the 34th ACM International Conference on Multimedia (MM '26), November 10--14, 2026, Rio de Janeiro, Brazil}
\acmDOI{10.1145/3767308.3836246}
\acmISBN{979-8-4007-2213-4/2026/11}

\usepackage{multirow}
\usepackage{pifont}
\usepackage{xspace}
\usepackage{enumitem}
\usepackage{booktabs}
\usepackage{tabularx}
\usepackage{float}

\usepackage{amssymb}

\newcommand{\model}{VTInstructor\xspace}

\begin{document}

\title[VTInstructor]{VTInstructor: Visual Trajectory Prompting for Navigation Instruction Generation in Continuous Environments}

\author{Haolin Yang}
\authornote{These authors contributed equally to this work.}
\affiliation{%
  \institution{CFCS, School of Computer Science, Peking University}
  \city{Beijing}
  \country{China}}
\affiliation{%
  \institution{PrimeBot}
  \city{Beijing}
  \country{China}}
\email{harley\_yang@stu.pku.edu.cn}

\author{Yuxing Long}
\authornotemark[1]
\affiliation{%
  \institution{CFCS, School of Computer Science, Peking University}
  \city{Beijing}
  \country{China}}
\affiliation{%
  \institution{PrimeBot}
  \city{Beijing}
  \country{China}}
\email{longyuxing@stu.pku.edu.cn}

\author{Zihan Yang}
\authornotemark[1]
\affiliation{%
  \institution{CFCS, School of Computer Science, Peking University}
  \city{Beijing}
  \country{China}}
\affiliation{%
  \institution{PrimeBot}
  \city{Beijing}
  \country{China}}
\email{yangzihan@stu.pku.edu.cn}

\author{Hao Dong}
\authornote{Corresponding author.}
\affiliation{%
  \institution{CFCS, School of Computer Science, Peking University}
  \city{Beijing}
  \country{China}}
\affiliation{%
  \institution{PrimeBot}
  \city{Beijing}
  \country{China}}
\email{hao.dong@pku.edu.cn}

\begin{abstract}
Navigation instruction generation from ego-centric RGB video in continuous environments is an important yet challenging task for human--robot interaction and scalable dataset construction.
Prior instruction generators assume discrete viewpoint graphs with panoramic observations, where trajectory structure is explicit; in continuous environments, however, the agent receives only a dense RGB stream, making trajectory cues difficult to recover.
We propose \textbf{\model}, the first VLN instruction generation framework for continuous environments. Our key idea is to convert implicit trajectory geometry into explicit visual trajectory prompts: EDTC condenses long RGB trajectories into navigation-critical keyframes, VTP overlays path, turn, and goal cues onto these anchors, VTMod injects the resulting trajectory signals into the visual encoder, and VT-GRPO further calibrates this spatial injection during training, all without requiring a navigation graph, pre-built map, or scene reconstruction.
On the challenging R2R-CE and RxR-CE \textit{Val Unseen} benchmarks, \model sets a new state of the art across all standard NLG metrics, surpassing the strongest baseline by +0.357 CIDEr and +0.109 CIDEr, respectively. Beyond automatic metrics, \model-generated instructions raise a frozen follower's success rate to 63.3\%, a +14.7 percentage-point gain over the best competing instruction source, and provide consistent data augmentation gains of +3 SR points on downstream navigation tasks.

\end{abstract}

\ccsdesc[500]{Computing methodologies~Vision for robotics}
\ccsdesc[300]{Computing methodologies~Natural language generation}
\ccsdesc[500]{Computing methodologies~Robotic planning}

\keywords{visual trajectory, multimodal robotics, computer vision}

\begin{teaserfigure}
  \centering
  \includegraphics[width=0.89\linewidth]{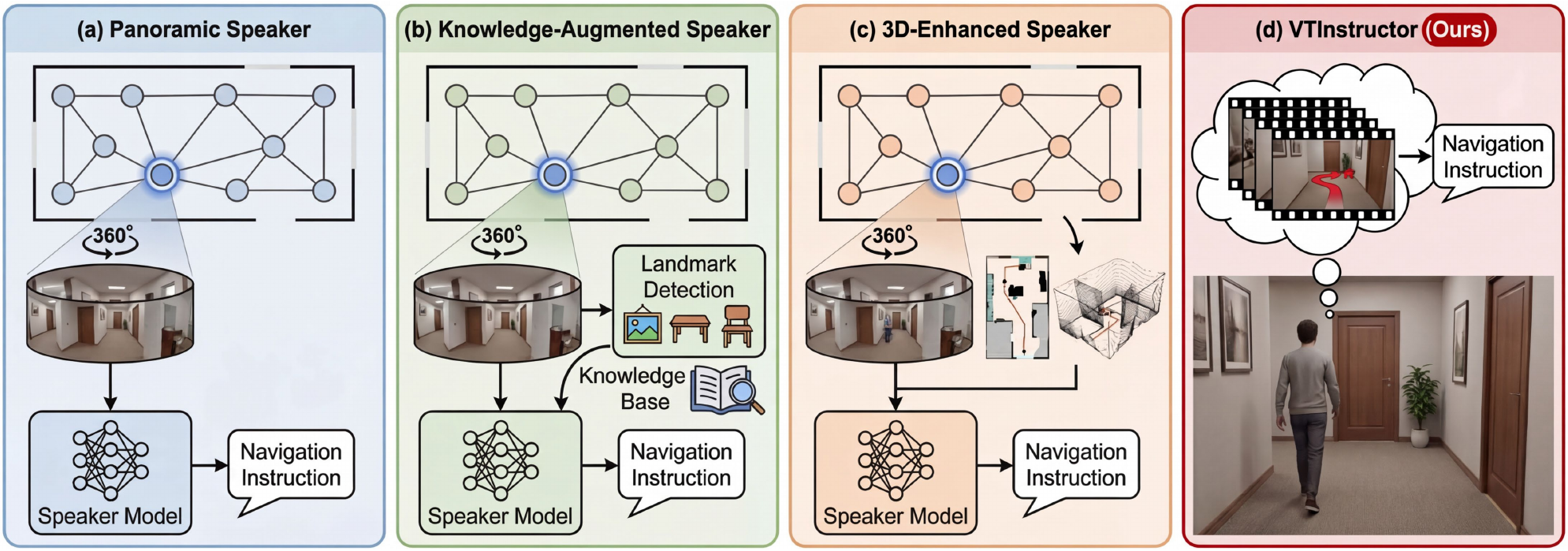}
  \caption{Evolution of navigation instruction generation paradigms. (a)–(c) Prior speakers rely on discrete viewpoint graphs with panoramic images, optionally augmented by external knowledge or 3D representations. (d) \model generates instructions from ego-centric RGB video in continuous environments, requiring no navigation graph or 3D reconstruction at inference.}
  \label{fig:teaser}
\end{teaserfigure}

\maketitle

\section{Introduction}
\vspace{-2pt}

Navigation instruction generation (producing natural-language descriptions of traversed trajectories) is a fundamental task for human-robot interaction and the scalable construction of training data for embodied navigation models.
Existing speaker models~\cite{fried2018sas, kamath2022newpath, li2022envedit} have made substantial progress in discrete graph-based settings. Figure~\ref{fig:teaser} summarizes the evolution of these paradigms, from panoramic speakers to knowledge-augmented and 3D-enhanced variants, all of which remain rooted in discrete viewpoint graphs. By contrast, navigation instruction generation in continuous environments remains largely unexplored.

Why has this setting remained unexplored? In discrete VLN environments, trajectories unfold over topological graphs whose nodes are associated with panoramic observations, making viewpoint relations explicit and trajectory structure easy to infer. In continuous environments, however, the agent receives only a dense ego-centric RGB stream, and adjacent frames may look highly similar even when they correspond to different motions or spatial states. This shift introduces two challenges. First, existing graph-based speaker models cannot be transferred directly to continuous environments. Second, trajectory cues such as path direction, turning behaviour, and goal progress are no longer explicit: recovering them from subtle inter-frame changes is itself a fine-grained spatial-intelligence problem, and this is precisely the capability that current VLN agents and MLLMs have been shown to lack~\cite{yang2026navspacenavigationagentsfollow}. Continuous environments therefore place instruction
generation squarely in the regime where these models are weakest, so
directional information is lost and instructions degenerate into generic, spatially imprecise descriptions---unless the trajectory geometry is supplied explicitly.

To address these challenges, we propose \textbf{\model}, a unified navigation instruction generation framework for continuous environments that processes the entire pipeline within a single vision-language backbone. Our central idea is to convert trajectory geometry into explicit visual trajectory prompts on ego-centric views, so that the model can perceive path structure directly instead of inferring it only from dense RGB streams. \model comprises three core components. (1)~\textit{Event-driven trajectory compression (EDTC)} first condenses long RGB trajectories into navigation-critical keyframes determined by the action sequence, providing visual anchors for subsequent prompting. (2)~\textit{Visual Trajectory Prompt (VTP) rendering} then overlays path, turn, and goal cues onto these keyframes, while GPT-based quality filtering (QF) retains only spatially reliable instruction-keyframe pairs for training. (3)~\textit{Visual Trajectory Modulator (VTMod) injection} feeds the resulting trajectory signals directly into the ViT encoder, strengthening spatial perception beyond what raw appearance alone can provide. On top of supervised fine-tuning, we further introduce VT-GRPO, which uses reinforcement learning to selectively calibrate the VTMod gates and explicitly refine how trajectory information is injected into the model.

We evaluate \model on the R2R-CE and RxR-CE \textit{Val Unseen} benchmarks, where it achieves state-of-the-art performance across all standard NLG metrics (BLEU, METEOR, ROUGE-L, CIDEr, SPICE), outperforming the strongest baseline. Downstream navigation experiments further show that follower agents guided by \model-generated instructions achieve higher success rate than those using competing instruction sources, and human evaluators consistently rate \model instructions higher on directional accuracy and overall followability.

Our contributions are:
\begin{itemize}[leftmargin=*,nosep]
    \item \textbf{The first VLN instruction generation framework for continuous environments.}
    \model generates instructions from ego-centric RGB trajectories paired with action sequences (as defined in Section~3.1); the model itself receives only RGB frames as visual input, without navigation graphs, pre-built maps, or scene reconstruction.

    \item \textbf{A visual trajectory prompting framework for explicit spatial grounding.}
    We convert implicit trajectory geometry in dense RGB streams into explicit spatial cues through EDTC for navigation-critical keyframe selection, VTP for path/turn/goal prompting on these anchors, VTMod for trajectory-aware visual encoding, and VT-GRPO for reward-driven calibration of spatial signal injection.

    \item \textbf{State-of-the-art performance with practical utility.}
    \model achieves state-of-the-art results on the R2R-CE and RxR-CE \textit{Val Unseen} benchmarks, surpassing the strongest baseline by +0.357 and +0.109 CIDEr, respectively, improving frozen-follower success by 14.7 percentage points, and delivering +3 SR-point data augmentation gains on both benchmarks.
\end{itemize}

\section{Related Work}
\vspace{-2pt}

\subsection{Navigation Instruction Generation}
\vspace{-2pt}

Vision-and-Language Navigation (VLN) requires an agent to follow natural-language instructions in indoor environments~\cite{anderson2018r2r, qi2020reverie, chen2022hamt}.
Complementing instruction \textit{following}, instruction \textit{generation} (the speaker side) is critical for data augmentation and human--robot communication.
Speaker-Follower~\cite{fried2018sas} first trains an LSTM speaker for data augmentation; subsequent work improves generation through speaker--follower cycle consistency~\cite{wang2022ccc} and multi-task joint training~\cite{wang2023lana}.
Another line enriches the speaker with external knowledge or landmark grounding: SAS~\cite{gopinathan2024sas} and KEFA~\cite{zeng2023kefa} introduce object-spatial attention and commonsense alignment, while landmark-based instruction generation methods such as Less is More~\cite{wang2022less} abstracts the trajectory into detected landmarks and generates instructions via a text-to-text model.
More recently, C-Instructor~\cite{kong2024cinstructor} adopts chain-of-thought prompting with a multimodal LLM, and MapInstructor~\cite{fan2025mapinstructor} and BEVInstructor~\cite{fan2024bevinstructor} leverage top-down maps or BEV representations for global spatial reasoning.
Despite their diversity, existing instruction-generation methods are developed in discrete navigation graphs, where each state is typically represented by privileged panoramic observations—often instantiated in R2R-style VLN as a set of 36 discretized views per viewpoint—rather than raw first-person continuous perception.
R2R-CE~\cite{krantz2020r2rce} and RxR-CE~\cite{ku2020rxr} extend VLN benchmarks to the continuous Habitat simulator, yet speaker models for this setting remain unexplored.
\model is the first VLN instruction generation framework for continuous environments, taking ego-centric RGB image with rendered VTP as input without graph topology privilege.

\vspace{-2pt}
\subsection{Visual Prompting}
\vspace{-2pt}

Visual prompting augments input images with task-relevant annotations to steer model attention without modifying model weights.
In the general vision-language domain, coloured circles~\cite{shtedritski2023whatdoes}, numbered markers~\cite{yang2023setofmark}, and iteratively overlaid arrows and keypoints~\cite{nasiriany2024pivot} have been used for referring expression comprehension, visual grounding, and action prediction.
In robotic manipulation, TraceVLA~\cite{zheng2025tracevla} overlays end-effector trajectory traces on the current observation to enhance spatial-temporal awareness of VLA policies, and Robotic Visual Instruction~\cite{li2025rovi} annotates images with visual cues to guide manipulation actions.
Despite the growing adoption of visual prompting in perception and manipulation, it has not been explored for navigation instruction generation.
\model is the first to introduce Visual Trajectory Prompts (VTP) into this task, with two key designs.

\section{Method}
\vspace{-2pt}

As shown in Figure 2, \model converts implicit trajectory geometry in dense ego-centric RGB streams into explicit spatial cues through four components: event-driven trajectory compression (§3.3), VTP rendering and data curation (§3.4), VTMod injection (§3.5), and VT-GRPO calibration (§3.6).

\begin{figure*}[t]
  \centering
  \includegraphics[width=0.89\linewidth]{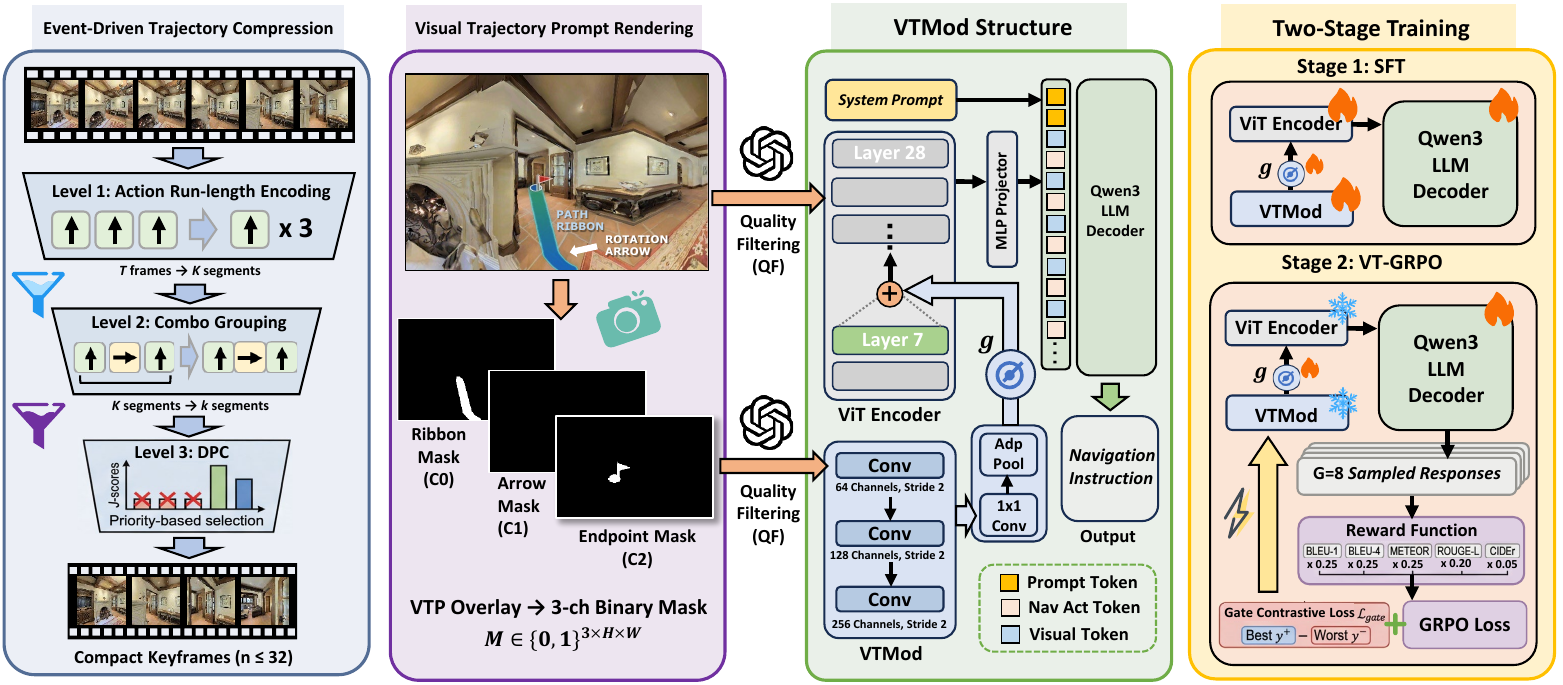}
  \caption{Overview of \model. Event-driven compression turns a raw ego-centric RGB trajectory into a compact keyframe set; each keyframe carries a VTP overlay (path ribbon / rotation arrow / endpoint marker) encoded as a 3-channel binary mask. The VTP Encoder maps the mask to patch-aligned features that are injected into ViT layer 7 by per-token spatial modulation. Training proceeds in two stages: SFT, then VT-GRPO.}
  \label{fig:overview}
\end{figure*}

\subsection{Problem Formulation}
\vspace{-2pt}
\label{sec:problem}

An agent navigates a 3D scene by executing atomic actions from a discrete action space $\mathcal{A}$ of four primitives: forward $+0.25\,\text{m}$, turn left $-30^{\circ}$, turn right $+30^{\circ}$, and stop.
At each timestep $t$, the agent receives an ego-centric RGB observation $f_t \in \mathbb{R}^{H\times W\times 3}$.
A trajectory of length $T$ is thus:
\begin{equation}
  \tau = \bigl((f_1, a_1),\, (f_2, a_2),\, \ldots,\, (f_T, a_T)\bigr).
\end{equation}

The \textit{Navigation Instruction Generation} (NIG) task requires producing a natural-language instruction $y = (w_1, \ldots, w_L)$ that describes $\tau$ faithfully enough for a human or autonomous follower to reproduce the route:
\begin{equation}
  P(y \mid \tau) = \prod_{l=1}^{L} P\!\left(w_l \;\middle|\; w_{<l},\, \tau\right).
\end{equation}

\vspace{-2pt}
\subsection{Preliminaries}
\vspace{-2pt}
\label{sec:base}

\noindent\textbf{Backbone Model.}
We build on \textbf{Qwen3-VL-8B}~\cite{qwen3vl}.
Its ViT-based visual encoder partitions each input image into non-overlapping patches forming a spatial token grid of shape $(h_p \times w_p)$ with hidden dimension $d_{\text{vit}}$; critically, each patch token $i$ retains a fixed spatial position $(r_i, c_i)$ throughout all layers, enabling the pixel-accurate one-to-one correspondence that VTMod exploits.
For $N$ keyframes, each frame is encoded independently and projected to the language model's hidden dimension by a shared MLP connector, with per-image 2D RoPE encodings preserving spatial and temporal order. On the language side, the system prompt and interleaved action snippets are tokenized into decoder input embeddings and combined with the projected visual tokens in a single multimodal sequence. The decoder then generates the navigation instruction autoregressively.

\noindent\textbf{Multimodal input format.}
The model receives a single interleaved sequence: a task-specific system prompt~$\mathcal{P}$ is followed by alternating keyframe images (I) and textual action snippets:
\[
  \mathcal{P},\; I_1,\; [\text{Action 1: } d_1],\; I_2,\; [\text{Action 2: } d_2],\; \ldots
\]
where each $d_k$ is a natural-language description of the physical displacement or rotation at that step (e.g.\ ``go forward 1.25\,m'', ``turn left 90$^{\circ}$'').
This interleaved layout preserves the temporal alignment between visual observations and physical actions, allowing the decoder to attend to the relevant image--action pair at each generation step.

\vspace{-2pt}
\subsection{Event-Driven Trajectory Compression}
\vspace{-2pt}
\label{sec:compression}

\noindent\textbf{Level 1: Action Run-Length Encoding (RLE).}
Let the raw action sequence be $\mathbf{a} = (a_1, a_2, \ldots, a_T)$ with $a_t \in \mathcal{A}$.
RLE merges consecutive identical actions into \textit{segments}:
\begin{equation}
  s_k = (\mathrm{type}_k,\ n_k), \quad \text{where } a_t = \mathrm{type}_k \text{ for } n_k \text{ consecutive steps.}
\end{equation}
Each segment maps to a physical displacement ($n_k \times 0.25\,\text{m}$ forward) or rotation ($n_k \times 30^{\circ}$), producing the compressed sequence $\mathbf{S} = (s_1, \ldots, s_K)$ with $K \ll T$.
RLE is lossless with respect to navigational semantics: every physical displacement and rotation is exactly preserved in $\mathbf{S}$.
To avoid overly long straight-line events, we further split any forward segment whose displacement exceeds $4.0\,\mathrm{m}$ (i.e., $n_k > 16$) into two shorter forward segments of approximately equal length. This preserves the total displacement while preventing excessively long forward motion from dominating a single event representation.

\noindent\textbf{Level 2: Small-Step Combo Grouping.}
A segment $s_k$ is considered \textit{small-step} if it corresponds to either a short forward displacement of $0.25\,\mathrm{m}$ or $0.5\,\mathrm{m}$ ($n_k \in \{1,2\}$) or a single-step rotation of $30^{\circ}$ ($n_k = 1$).
Consecutive small-step segments are merged into a \textit{combo event}
\[
e_c = (s_k, s_{k+1}, \ldots, s_{k+M-1}),
\]
with $M \leq 6$.
This grouping captures fine-grained turn-and-advance manoeuvres as a single semantic unit, preventing them from being fragmented into isolated snippets that may appear spatially incoherent to the language model.

\noindent\textbf{Level 3: Dynamic Priority-based Compression (DPC).}
When the number of retained frames after Levels 1--2 still exceeds $F_{\max} = 32$, DPC selects the most informative keyframes by a joint visual--geometric priority score.
For each adjacent event-frame pair $(i, j)$:
\begin{equation}
  J(i,j) = \lambda_{\text{vis}}\,\Delta_{\text{vis}}(i,j) + \lambda_{\text{geo}}\,\Delta_{\text{geo}}(i,j),
\end{equation}
where $\Delta_{\text{vis}}$ measures frame-to-frame visual change in a frozen perceptual feature space, and $\Delta_{\text{geo}}$ aggregates the cumulative displacement and rotation magnitude between event boundaries.
A high $J$ score indicates a visually or geometrically significant transition that should be retained; event frames at boundaries with the \emph{lowest} $J$ scores are progressively removed until $|\mathcal{F}| \leq F_{\max}$, while always preserving the initial frame and the current observation frame.
The final retained frame set is:
\begin{equation}
  \mathcal{F} = \{f_1, f_T\} \cup \{f_{e_1}, f_{e_2}, \ldots, f_{e_{N'}}\},
\end{equation}
where $f_1$ is the initial observation, $f_T$ is the current observation frame, $e_i$ denotes the index of the $i$-th retained event boundary, and $N'$ is the number of retained event frames.

\vspace{-2pt}
\subsection{Visual Trajectory Prompt Rendering and Data Curation}
\vspace{-2pt}
\label{sec:vp_rendering}

\noindent\textbf{Overlay components.}
For each retained keyframe, a structured VTP is rendered in the corresponding ego-centric view to visualize the local trajectory state.
Three complementary components provide exhaustive spatial coverage across the canonical navigation states (advancing, turning, approaching goal):

\begin{itemize}[leftmargin=*,nosep]
  \item \textbf{Path ribbon}: a colour ribbon tracing the upcoming route in the current view. Because the raw path is a jagged polyline induced by $0.25\,\mathrm{m}$ atomic forward steps, we smooth it in 2D through densified interpolation, Chaikin corner cutting, and two rounds of Gaussian smoothing. A final occlusion check is then applied to prevent smoothed points from drifting behind walls. Used when the current event is not a pure rotation; degrades gracefully to a short stub when fully occluded.
  \item \textbf{Rotation arrow}: a curved left/right arc annotated with the turn angle in degrees. Used when the current event is a pure rotation.
  \item \textbf{Endpoint marker}: a landmark flag indicating the goal location in the current view. Rendered near trajectory end-frames when the goal is unoccluded.
\end{itemize}

\noindent\textbf{Mask representation.}
Each VTP overlay is stored as a three-channel binary semantic mask $\mathbf{M} \in \{0,1\}^{3 \times H \times W}$, where channel $C_0$ encodes the ribbon, $C_1$ the rotation arrow, and $C_2$ the endpoint marker.
Representing overlays as independent binary channels eliminates colour ambiguity and lets the VTP Encoder learn channel-specific spatial patterns, a factorisation that would be conflated in a mixed-colour image.
This mask is the direct input to the VTP Encoder; the coloured PNG visualisation is an artefact used only for qualitative inspection.

\noindent\textbf{Task-specific prompting strategy.}
We design separate system prompts for R2R-CE and RxR-CE to match their distinct annotation styles: the R2R-CE prompt targets concise instructions (15--45 words) emphasising landmark references and a precise stop location, while the RxR-CE prompt elicits richer step-by-step narrations (50--120 words) with explicit orientation and transition cues.
Both prompts share a critical \textit{use-but-don't-mention} constraint: the model is informed that the input images contain grounded navigation cues (path ribbons, turn indicators, goal markers) and is instructed to leverage these overlays for path inference, yet is explicitly prohibited from referencing them in the generated instruction.
This design ensures VTP functions as an implicit geometric prior that improves spatial grounding without leaking rendering artefacts into the output text.

\noindent\textbf{GPT-based quality filtering.}
Instruction candidates are scored by GPT against a rubric covering four dimensions: (i)~\textit{directional accuracy}: do described turns and path shape match the VTP overlay?; (ii)~\textit{landmark specificity}: are salient visual features referenced?; (iii)~\textit{distance plausibility}: do distance expressions correspond to actual trajectory length?; and (iv)~\textit{linguistic fluency}.
Only samples exceeding quality threshold $\tau_{\text{GPT}}$ are retained, removing hallucinated or spatially imprecise instructions and ensuring the training corpus maintains consistent geometric fidelity.

\vspace{-2pt}
\subsection{Visual Trajectory Modulator Design}
\vspace{-2pt}
\label{sec:vtmod}

\noindent\textbf{VTP Encoder.}
Let $\mathbf{M} \in \{0,1\}^{3 \times H \times W}$ denote the binary semantic mask with $\mathbf{F}^{(0)} = \mathbf{M}$.
The VTP Encoder applies three stride-2 convolutional blocks, each comprising a convolution, group normalisation, and GELU activation, with output channel dimensions $C_l \in \{64, 128, 256\}$:
\begin{equation}
  \mathbf{F}^{(l)} = \mathrm{GELU}\!\left(\mathrm{GN}\!\left(\mathrm{Conv}_{s=2}^{(l)}\bigl(\mathbf{F}^{(l-1)}\bigr)\right)\right), \quad l = 1,2,3.
\end{equation}
The hierarchical stride-2 design progressively expands the receptive field so that each spatial position encodes not only its local overlay pixel but also the surrounding trajectory context.
A subsequent $1{\times}1$ convolution projects to dimension $d_v = 384$, and adaptive average pooling aligns the spatial resolution to the ViT patch grid, yielding:
\begin{equation}
  \mathbf{V} = \mathrm{AdpPool}\!\left(\mathrm{Conv}_{1\times 1}\bigl(\mathbf{F}^{(3)}\bigr),\,(h_p, w_p)\right) \in \mathbb{R}^{h_p w_p \times d_v},
\end{equation}
where each row $\mathbf{v}_i \in \mathbb{R}^{d_v}$ is in one-to-one spatial correspondence with the $i$-th ViT patch token.

\noindent\textbf{Spatial modulation injection.}
At ViT layer $l^{*} = 7$, the hidden state $\mathbf{h}_i \in \mathbb{R}^{d_{\mathrm{vit}}}$ of patch token $i$ is updated as:
\begin{equation}
  \mathbf{h}'_i = \mathbf{h}_i + \mathbf{g} \odot \mathrm{LN}(\mathbf{W}\mathbf{v}_i),
\end{equation}
where $\mathbf{W} \in \mathbb{R}^{d_{\mathrm{vit}} \times d_v}$ is a learnable linear projection, $\mathrm{LN}$ denotes layer normalisation, $\odot$ denotes element-wise multiplication, and $\mathbf{g} \in \mathbb{R}^{d_{\mathrm{vit}}}$ is a channel-wise learnable gate.
We set $l^{*} = 7$ as the default injection site (ablated in  Table~\ref{tab:ablation_design}).
The gate $\mathbf{g}$ is initialised to $\mathbf{g}_0 \approx \mathbf{0}$, implementing a \textit{lazy activation}: the pre-trained ViT gradient landscape is preserved at training onset, preventing catastrophic forgetting of visual priors while the backbone adapts to the new task during early SFT.

\vspace{-2pt}
\subsection{Training Strategy}
\vspace{-2pt}
\label{sec:training}

\noindent\textbf{Two-stage rationale.}
The two stages address complementary limitations.
SFT teaches the task distribution (what high-quality navigation instructions conditioned on VTP-annotated keyframes look like) via maximum-likelihood estimation on GPT-filtered data.
However, MLE treats all reference tokens equally regardless of navigational informativeness, and cannot directly optimise full-sequence metrics such as CIDEr or METEOR that are evaluated at test time.
VT-GRPO then uses NLG reward signals to calibrate the gate $\mathbf{g}$'s injection strength while continuing to update the LLM backbone, sharpening which spatial channels of the VTP are amplified for reward-relevant generation.

\noindent\textbf{Stage 1: Supervised Fine-Tuning (SFT).}
During SFT, all modules are trainable, including the ViT backbone, VTP Encoder, VTMod, MLP connector, and Qwen3 decoder. We use a dual learning-rate scheme: a higher rate $\eta_{\text{new}}$ for newly initialised VTP/VTMod parameters and a lower rate $\eta_{\text{backbone}}$ for the pretrained Qwen3-VL backbone.
The SFT objective is standard next-token prediction cross-entropy over ground-truth instruction tokens.

\noindent\textbf{Stage 2: VT-GRPO.}\
Group Relative Policy Optimisation (GRPO)~\cite{shao2024deepseekmath} is applied to refine instruction quality using NLG metrics as reward.
For each input, $G = 8$ candidate completions are sampled.
The reward for completion $y$ is a weighted combination of automatic metrics:
\begin{equation}
  \begin{split}
    r(y) = {}&w_{B1}\,\mathrm{B\text{-}1}(y) + w_{B4}\,\mathrm{B\text{-}4}(y) + w_M\,\mathrm{METEOR}(y)\\
              &+ w_R\,\mathrm{ROUGE\text{-}L}(y) + w_C\,\mathrm{CIDEr}(y),
  \end{split}
\end{equation}
with $w_{B-1} = w_{B-4} = w_M = 0.25$, $w_R = 0.20$, $w_C = 0.05$.
The weights reflect the complementary coverage of the metrics: B-1 (BLEU-1) captures unigram precision, B-4 (BLEU-4) rewards multi-word phrase fidelity, METEOR additionally accounts for synonym overlap, and ROUGE-L measures longest-common-subsequence structural similarity; CIDEr is down-weighted because its large absolute scale would otherwise dominate the composite reward.
A KL penalty $\beta = 0.04$ and clip ratio $\varepsilon = 0.2$ regularise the policy update.
Let $\mathcal{L}_{\text{GRPO}}$ denote the corresponding GRPO objective induced by these rewards and regularizers.

\noindent\textbf{Gate contrastive loss.}
Because $\mathcal{L}_{\text{GRPO}}$ averages over all $G$ completions weighted by their respective advantages, positive and negative signals partially cancel, leaving the gate $\mathbf{g}$ with a diffuse gradient that is insufficient for precise calibration.
We therefore introduce a contrastive loss on the best ($y^+$) and worst ($y^-$) completions within each group:
\begin{equation}
  \mathcal{L}_{\text{gate}} = -\log\,\sigma\!\left(\frac{\bar{\ell}(y^+) - \bar{\ell}(y^-)}{\tau_g}\right),
\end{equation}
where $\bar{\ell}(y)$ is the mean per-token log-probability and $\tau_g = 1.0$.
Unlike the group-averaged GRPO signal, this loss provides a focused contrastive gradient that directly pushes $\mathbf{g}$ to amplify VTP channels correlated with higher-reward generations and suppress those correlated with lower-reward ones.
The total training objective is:
\begin{equation}
  \mathcal{L}_{\text{total}} = \mathcal{L}_{\text{GRPO}} + \alpha_{\text{gate}}\,\mathcal{L}_{\text{gate}}, \quad \alpha_{\text{gate}} = 0.05.
\end{equation}
During VT-GRPO, the ViT backbone, VTP Encoder, and modulator projection layers are frozen; the LLM backbone and the gate vector $\mathbf{g}$ remain trainable.
Detailed hyperparameters for both stages are reported in \S\ref{sec:expdetails}.

\section{Experiments}
\vspace{-2pt}

\begin{table*}[t]
\centering
\footnotesize
\caption{Navigation instruction generation results on R2R-CE and RxR-CE \textit{Val Unseen}.}
\label{tab:main}
\setlength{\tabcolsep}{4pt}
\scalebox{0.83}{
\begin{tabularx}{\linewidth}{Xcccccccccccc}
\toprule
& \multicolumn{6}{c}{\textbf{R2R-CE} \textit{Val Unseen}}
& \multicolumn{6}{c}{\textbf{RxR-CE} \textit{Val Unseen}} \\
\cmidrule(lr){2-7}\cmidrule(lr){8-13}
\textbf{Method}
  & \textbf{BLEU-1} & \textbf{BLEU-4} & \textbf{METEOR} & \textbf{ROUGE-L} & \textbf{CIDEr} & \textbf{SPICE}
  & \textbf{BLEU-1} & \textbf{BLEU-4} & \textbf{METEOR} & \textbf{ROUGE-L} & \textbf{CIDEr} & \textbf{SPICE} \\
\midrule
\multicolumn{13}{l}{\textit{Proprietary models}} \\[1pt]
\quad Qwen3.5-Plus
  & 0.557 & 0.131 & 0.223 & 0.387 & 0.137 & 0.180 & 0.527 & 0.080 & 0.157 & 0.254 & 0.033 & 0.156 \\
\quad GPT-5.4
  & 0.482 & 0.075 & 0.197 & 0.348 & 0.078 & 0.144 & 0.493 & 0.056 & 0.146 & 0.241 & 0.029 & 0.143 \\
\quad Gemini-3.1-Pro-Preview
  & 0.631 & 0.166 & 0.209 & 0.392 & 0.203 & 0.167 & 0.393 & 0.073 & 0.128 & 0.242 & 0.029 & 0.135 \\
\quad xAI Grok 4
  & 0.486 & 0.084 & 0.207 & 0.354 & 0.073 & 0.160 & 0.430 & 0.049 & 0.181 & 0.249 & 0.014 & 0.144 \\
\quad Claude Opus 4.6
  & 0.484 & 0.084 & 0.190 & 0.337 & 0.068 & 0.119 & 0.367 & 0.043 & 0.176 & 0.223 & 0.005 & 0.123 \\
\midrule
\multicolumn{13}{l}{\textit{Open-source models}} \\[1pt]
\quad Qwen3.5-397B-A17B
  & 0.525 & 0.113 & 0.216 & 0.382 & 0.120 & 0.174 & 0.422 & 0.059 & 0.128 & 0.249 & 0.027 & 0.142 \\
\quad Kimi-K2.5
  & 0.478 & 0.088 & 0.195 & 0.357 & 0.098 & 0.146 & 0.432 & 0.071 & 0.195 & 0.259 & 0.011 & 0.145 \\
\quad LLaVA-Video-7B-Qwen2
  & 0.506 & 0.100 & 0.146 & 0.317 & 0.108 & 0.109 & 0.064 & 0.012 & 0.063 & 0.169 & 0.002 & 0.074 \\
\quad Qwen3-VL-8B-Instruct
  & 0.524 & 0.085 & 0.175 & 0.334 & 0.147 & 0.137 & 0.366 & 0.053 & 0.119 & 0.230 & 0.022 & 0.113 \\
\quad Qwen3.5-9B
  & 0.579 & 0.121 & 0.194 & 0.373 & 0.181 & 0.151 & 0.515 & 0.077 & 0.171 & 0.259 & 0.027 & 0.164 \\
\quad GLM-4.1V-9B-Thinking
  & 0.490 & 0.080 & 0.156 & 0.321 & 0.130 & 0.115 & 0.074 & 0.014 & 0.054 & 0.144 & 0.003 & 0.064 \\
\quad GLM-4.6V
  & 0.556 & 0.120 & 0.171 & 0.348 & 0.168 & 0.122 & 0.489 & 0.065 & 0.143 & 0.242 & 0.027 & 0.145 \\
\midrule
\multicolumn{13}{l}{\textit{Ours}} \\[1pt]
\quad Qwen3-VL-8B (SFT-only)
  & 0.720 & 0.282 & 0.232 & 0.485 & 0.484 & 0.201 & 0.630 & 0.201 & 0.210 & 0.357 & 0.060 & 0.162 \\
\quad \model\
  & \textbf{0.765} & \textbf{0.320} & \textbf{0.263} & \textbf{0.511} & \textbf{0.560} & \textbf{0.245}
  & \textbf{0.774} & \textbf{0.308} & \textbf{0.265} & \textbf{0.431} & \textbf{0.142} & \textbf{0.206} \\
\bottomrule
\end{tabularx}
}
\end{table*}

\begin{table*}[t]
\centering
\footnotesize
\caption{Cross-setting comparison with prior instruction generation methods. Prior methods use discretized viewpoint graphs with privileged 36-view observations; \model uses only ego-centric RGB in continuous environments. We present this as a reference point for instruction quality, not a strictly controlled comparison.}
\label{tab:cross_setting}
\setlength{\tabcolsep}{4pt}
\scalebox{0.83}{
\begin{tabularx}{\linewidth}{Xcccccccccccc}
\toprule
& \multicolumn{6}{c}{\textbf{R2R / R2R-CE} \textit{Val Unseen}}
& \multicolumn{6}{c}{\textbf{RxR / RxR-CE} \textit{Val Unseen}} \\
\cmidrule(lr){2-7}\cmidrule(lr){8-13}
\textbf{Method}
  & \textbf{BLEU-1} & \textbf{BLEU-4} & \textbf{METEOR} & \textbf{ROUGE-L} & \textbf{CIDEr} & \textbf{SPICE}
  & \textbf{BLEU-1} & \textbf{BLEU-4} & \textbf{METEOR} & \textbf{ROUGE-L} & \textbf{CIDEr} & \textbf{SPICE} \\
\midrule
\multicolumn{13}{l}{\textit{Prior methods on R2R / RxR (discretized viewpoint-graph environments)}} \\[1pt]
\quad BT-speaker~\cite{fried2018sas}
  & 0.658 & 0.250 & 0.209 & 0.440 & 0.391 & 0.178
  & 0.311 & 0.064 & 0.160 & 0.243 & 0.022 & -- \\
\quad EDrop-speaker~\cite{tan2019envdrop}
  & 0.660 & 0.260 & 0.215 & 0.455 & 0.413 & 0.184
  & 0.294 & 0.056 & 0.143 & 0.253 & 0.027 & -- \\
\quad CCC-speaker~\cite{wang2022ccc}
  & 0.679 & 0.254 & 0.226 & 0.456 & 0.401 & 0.183
  & 0.266 & 0.042 & 0.113 & 0.206 & 0.016 & -- \\
\quad Lana~\cite{wang2023lana}
  & 0.689 & 0.260 & 0.219 & 0.463 & 0.419 & 0.194
  & 0.314 & 0.111 & 0.126 & 0.267 & 0.045 & -- \\
\quad Lana+~\cite{wang2023lana}
  & 0.692 & 0.262 & 0.220 & 0.462 & 0.424 & 0.199
  & 0.308 & 0.115 & 0.125 & 0.268 & 0.046 & -- \\
\quad C-Instructor~\cite{kong2024cinstructor}
  & 0.711 & 0.263 & 0.237 & 0.469 & 0.450 & 0.211
  & 0.366 & 0.124 & 0.181 & 0.301 & 0.053 & -- \\
\quad BEVInstructor~\cite{fan2024bevinstructor}
  & 0.699 & 0.264 & 0.230 & 0.467 & 0.449 & 0.208
  & 0.351 & 0.112 & 0.174 & 0.292 & 0.043 & -- \\
\quad MapInstructor~\cite{fan2025mapinstructor}
  & 0.715 & 0.285 & 0.234 & 0.485 & 0.490 & 0.209
  & 0.411 & 0.159 & 0.191 & 0.323 & 0.057 & -- \\
\midrule
\multicolumn{13}{l}{\textit{Ours on R2R-CE / RxR-CE (continuous environments)}} \\[1pt]
\quad \model\ 
  & \textbf{0.765} & \textbf{0.320} & \textbf{0.263} & \textbf{0.511} & \textbf{0.560} & \textbf{0.245}
  & \textbf{0.774} & \textbf{0.308} & \textbf{0.265} & \textbf{0.431} & \textbf{0.142} & \textbf{0.206} \\
\bottomrule
\end{tabularx}
}
\end{table*}

\subsection{Implementation Details}
\vspace{-2pt}
\label{sec:expdetails}

\noindent\textbf{Hardware.}
All experiments are conducted on 8$\times$NVIDIA H200 GPUs with DeepSpeed ZeRO-2 parallelism.

\noindent\textbf{Training data.}
Training data consists of GPT-quality-filtered VTP-annotated trajectories (score $\geq 6$ on a 10-point rubric) drawn exclusively from the Train splits of R2R-CE and RxR-CE.

\noindent\textbf{Input representation.}
Each observation is represented as a single wide-aspect egocentric image, with VTP overlays rendered only in the central region.
Trajectories are compressed to at most $F_{\max} = 32$ keyframes using the event-driven pipeline (\S\ref{sec:compression}).

\noindent\textbf{SFT hyperparameters.}
3 epochs; batch size 1; gradient accumulation 12; base LR $3 \times 10^{-5}$ (backbone), VTMod LR $5 \times 10^{-4}$; 10\% linear warm-up. Training takes approximately 9 hours on the above hardware.

\noindent\textbf{VT-GRPO hyperparameters.}
1 epoch; LR $10^{-6}$; gradient accumulation 4; group size $G = 8$; KL $\beta = 0.04$; clip $\varepsilon = 0.2$; top-$p = 0.9$; gate-contrastive weight $\alpha_{\text{gate}} = 0.05$. Training takes approximately 30 hours on the above hardware.

\vspace{-2pt}
\subsection{Evaluation Benchmarks and Metrics}
\vspace{-2pt}

\noindent\textbf{R2R-CE}~\cite{krantz2020r2rce} extends the Room-to-Room benchmark~\cite{anderson2018r2r} to the continuous-action Habitat simulator with photorealistic Matterport3D (MP3D) scenes.
The agent navigates via four atomic actions and receives only a raw ego-centric RGB stream, with no navigation graph or pre-built map available.
The \textit{Val Unseen} split covers environments entirely held out from training, providing a stringent test of instruction generalisation to novel scenes.
Since \model is trained exclusively on the Train splits of R2R-CE and RxR-CE, neither the environments nor the reference instructions in \textit{Val Unseen} have been seen during training.

\noindent\textbf{RxR-CE}~\cite{ku2020rxr} adapts the Room-across-Room benchmark to the same Habitat continuous setting.
RxR instructions are substantially longer and more spatially detailed than R2R (averaging over 70 words per instruction), with fine-grained descriptions of turn angles, landmark sequences, and relative distances.
This verbosity and spatial precision make RxR-CE a more demanding benchmark for evaluating instruction generation quality.
We report results on the English \textit{Val Unseen} split.

\noindent\textbf{Instruction metrics.}
We report BLEU-1/4, METEOR, ROUGE-L, CIDEr, and SPICE; higher is better for all.
These metrics collectively capture $n$-gram precision, recall, synonym overlap, sequential similarity, and semantic propositional content.

\noindent\textbf{Navigation metrics.}
For downstream experiments (\S\ref{sec:downstream}), we report the standard VLN-CE metrics~\cite{anderson2018r2r}: Success Rate (SR), Oracle Success Rate (OSR), SPL, and Navigation Error (NE, in metres; lower is better), all with a 3\,m success threshold.

\vspace{-2pt}
\subsection{Main Results}
\vspace{-2pt}

\noindent\textbf{Comparison with existing models (Table~\ref{tab:main}).}
All baseline models are evaluated under a few-shot setting: each prompt includes a small number of high-quality reference instructions sampled from the training split that exemplify the target annotation style of R2R-CE or RxR-CE, ensuring that every model receives sufficient task context before generation.
\model surpasses all baselines by a large margin across both benchmarks.
On R2R-CE, \model achieves a CIDEr of 0.560, outperforming the strongest model Gemini-3.1-Pro-Preview (0.203) by +0.357; BLEU-4 reaches 0.320 versus 0.166 for Gemini, an improvement of over 90\%.
On RxR-CE, the gap widens further: most baselines obtain CIDEr below 0.03, while \model reaches 0.142.
Notably, LLaVA-Video-7B and GLM-4.1V-9B-Thinking nearly collapse on RxR-CE (BLEU-1 of 0.064 and 0.074, respectively), suggesting that these models struggle with the longer observation sequences required by RxR trajectories.
Notably, proprietary models do not consistently outperform smaller open-source ones---GPT-5.4 (CIDEr 0.078) falls below Qwen3.5-9B (0.181) on R2R-CE---indicating that navigation instruction generation cannot be solved by model scale alone and benefits from task-specific training with spatial grounding.

\begin{figure*}[t]
  \centering
  \includegraphics[width=0.88\linewidth]{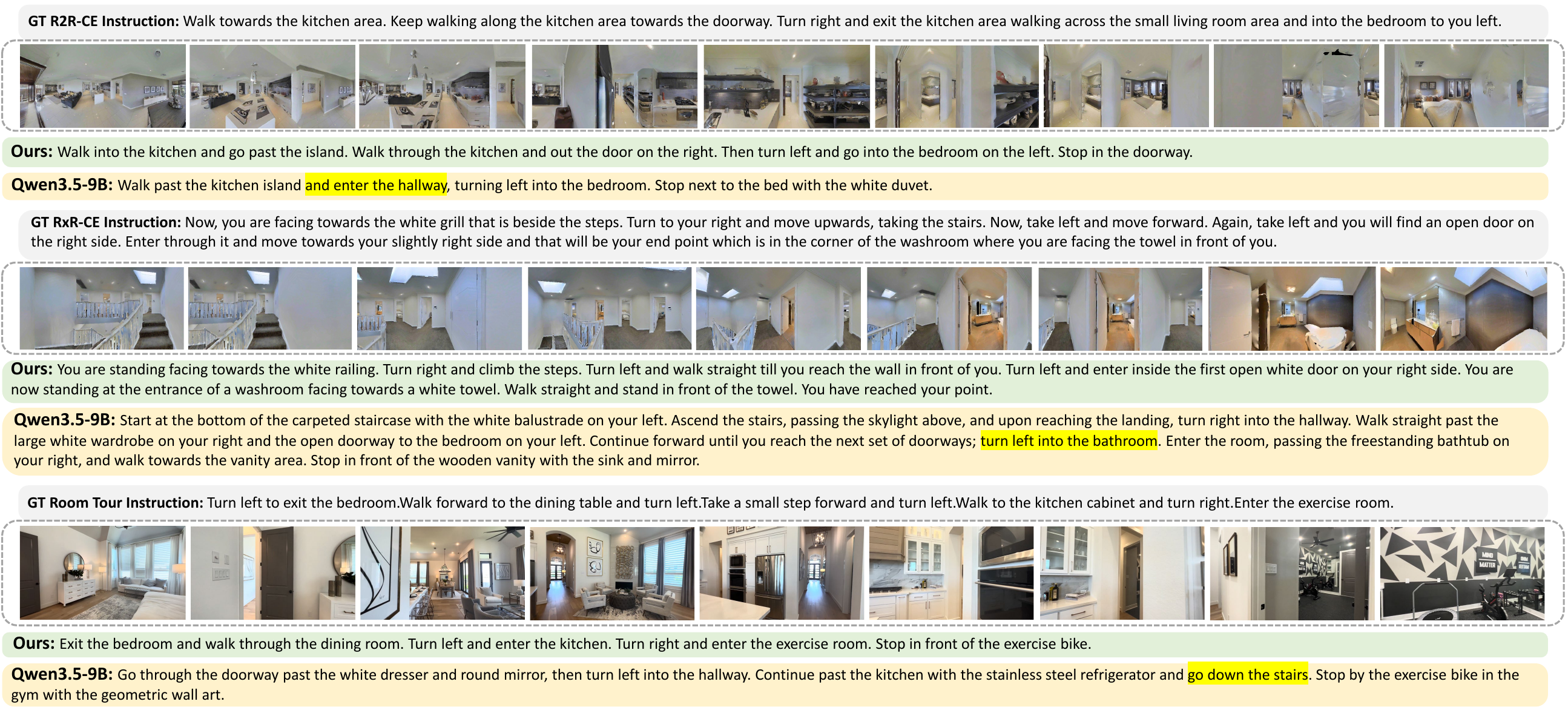}
  \caption{Qualitative comparison on R2R-CE and RxR-CE \textit{Val Unseen} and a real-world room-tour clip. Each case shows the compressed keyframes with VTP overlays (top) and instructions from the ground truth, the strongest baseline, and \model (bottom). Yellow highlights mark factual errors in the baseline output (e.g. wrong turn direction, hallucinated landmarks).}

  \label{fig:qualitative}
\end{figure*}

\noindent\textbf{Cross-setting comparison (Table~\ref{tab:cross_setting}).}
Operating in the continuous setting, \model attains scores comparable to or higher than prior discrete-setting methods that have access to panoramic images and pre-built navigation graphs. Because the two settings differ in observation and action spaces, we treat Table~\ref{tab:cross_setting} as a reference point for instruction quality rather than a head-to-head controlled comparison. On R2R, \model surpasses the previous best MapInstructor in BLEU-4 (0.320 vs.\ 0.285), METEOR (0.263 vs.\ 0.234), and CIDEr (0.560 vs.\ 0.490). The advantage is even more pronounced on RxR, where \model achieves BLEU-4 of 0.308 versus MapInstructor's 0.159 and CIDEr of 0.142 versus 0.057, suggesting that VTP-based spatial grounding is particularly effective for the longer, more spatially detailed instructions characteristic of RxR. R2R-CE/RxR-CE and R2R/RxR share identical paths and reference instructions, and we follow the per-sub-path aggregation protocol of C-Instructor~[8], so the metrics are comparable. Figure~\ref{fig:qualitative} provides qualitative examples illustrating how \model generates spatially accurate instructions while the strongest baseline produces directional errors and hallucinated landmarks.

\vspace{-2pt}
\subsection{Ablation Study}
\vspace{-2pt}

\begin{table}[t]
\centering
\footnotesize
\caption{Component-wise ablation on R2R-CE \textit{Val Unseen}.}
\label{tab:ablation}
\setlength{\tabcolsep}{4pt}
\renewcommand{\arraystretch}{1.05}
\scalebox{0.83}{
\begin{tabularx}{\columnwidth}{c c c c c c c c}
\toprule
\textbf{\#} & \textbf{VTMod} & \textbf{EDTC} & \textbf{QF} & \textbf{BLEU-4} & \textbf{METEOR} & \textbf{ROUGE-L} & \textbf{SPICE} \\
\midrule
1 &  &  &  & 0.282 & 0.232 & 0.485 & 0.201 \\
2 &  &  & $\checkmark$ & 0.286 & 0.241 & 0.492 & 0.220 \\
3 &  & $\checkmark$ & $\checkmark$ & 0.288 & 0.242 & 0.493 & 0.223 \\
4 & $\checkmark$ & $\checkmark$ &  & 0.302 & 0.250 & 0.500 & 0.232 \\
5 & $\checkmark$ & $\checkmark$ & $\checkmark$ & \textbf{0.308} & \textbf{0.256} & \textbf{0.504} & \textbf{0.238} \\
\bottomrule
\end{tabularx}
}
\end{table}

\begin{table}[t]
\centering
\footnotesize
\caption{Design ablations on R2R-CE \textit{Val Unseen}.}
\label{tab:ablation_design}
\setlength{\tabcolsep}{4pt}
\renewcommand{\arraystretch}{1.05}
\scalebox{0.83}{
\begin{tabularx}{\columnwidth}{X c c c c}
\toprule
\textbf{Configuration} & \textbf{BLEU-4} & \textbf{METEOR} & \textbf{ROUGE-L} & \textbf{SPICE} \\
\midrule
\multicolumn{5}{l}{\textit{(a) EDTC: frame sampling}} \\
\quad Stride-4 (avg.\ 13.0 fr.) & 0.738$^\dagger$ & 0.240 & 0.490 & 0.220 \\
\quad EDTC (avg.\ 10.5 fr.) & \textbf{0.743}$^\dagger$ & \textbf{0.242} & \textbf{0.493} & \textbf{0.223} \\
\midrule
\multicolumn{5}{l}{\textit{(b) VTMod: injection layer}} \\
\quad Layer 7 & \textbf{0.299} & \textbf{0.248} & \textbf{0.497} & \textbf{0.224} \\
\quad Layers 7, 15, 23 & 0.295 & 0.244 & 0.491 & 0.221 \\
\midrule
\multicolumn{5}{l}{\textit{(c) VTMod: injection design}} \\
\quad Cross-attention & 0.295 & 0.244 & 0.494 & 0.222 \\
\quad Gated projection & \textbf{0.300} & \textbf{0.247} & \textbf{0.497} & \textbf{0.226} \\
\midrule
\multicolumn{5}{l}{\textit{(d) VT-GRPO: RL strategy}} \\
\quad Pure GRPO & 0.314 & 0.260 & 0.508 & 0.240 \\
\quad VT-GRPO & \textbf{0.320} & \textbf{0.263} & \textbf{0.511} & \textbf{0.245} \\
\bottomrule
\multicolumn{5}{l}{\scriptsize $^\dagger$\,BLEU-1 reported for frame sampling comparison.} \\
\end{tabularx}
}
\end{table}

We conduct ablation experiments on R2R-CE \textit{Val Unseen}, whose shorter reference instructions make component contributions easier to isolate.

\noindent\textbf{Component-wise ablation.}
Table~\ref{tab:ablation} progressively adds VTMod, EDTC, and quality filtering (QF).
The base SFT model without any of these components (Row~1) achieves BLEU-4 of 0.282 and SPICE of 0.201.
Adding VTMod yields the largest single improvement: comparing Row~3 to Row~5, BLEU-4 jumps from 0.288 to 0.308 (+0.020) and SPICE from 0.223 to 0.238 (+0.015), confirming that VTMod is the core contributor by injecting trajectory-grounded spatial cues into the vision encoder.
The full model (Row~5) achieves BLEU-4 of 0.308, METEOR of 0.256, ROUGE-L of 0.504, and SPICE of 0.238, representing cumulative gains of +0.026, +0.024, +0.019, and +0.037 over the base model. These results are obtained solely with supervised fine-tuning (SFT), without applying the subsequent VT-GRPO optimization process.

\noindent\textbf{Design ablations.}
As shown in Table~\ref{tab:ablation_design}, EDTC reduces the average frame count from 13.0 to 10.5 while improving all metrics over stride-4 sampling, confirming that event-driven selection retains more informative keyframes.
Injecting VTMod at a single early layer (layer~7) outperforms distributing it across layers 7/15/23 (BLEU-4 0.299 vs.\ 0.295), as subsequent ViT layers can jointly refine the fused representation without redundant modulation.
Gated patchwise projection outperforms cross-attention injection (BLEU-4 0.300 vs.\ 0.295), preserving spatial locality of VTP cues that cross-attention would dilute.
Finally, VT-GRPO improves over pure GRPO (BLEU-4 0.314$\to$0.320, SPICE 0.240$\to$0.245), demonstrating that the gate contrastive reward yields more spatially grounded instructions.

\vspace{-2pt}
\subsection{Downstream Navigation Success Rate}
\vspace{-2pt}
\label{sec:downstream}

\begin{table}[t]
\centering
\small
\caption{Downstream navigation performance on R2R-CE \textit{Val Unseen} with a frozen CorrectNav follower.}
\label{tab:downstream}
\setlength{\tabcolsep}{5pt}
\scalebox{0.83}{
\begin{tabularx}{\columnwidth}{X@{\hspace{10pt}}cccc}
\toprule
\textbf{Instruction Source}
  & \textbf{SR}$\uparrow$
  & \textbf{OSR}$\uparrow$
  & \textbf{SPL}$\uparrow$
  & \textbf{NE}$\downarrow$ \\
\midrule
\multicolumn{5}{l}{\textit{Human Annotations}} \\[1pt]
\quad Official instruction            & 61.6 & 67.2 & 53.3 & 4.53 \\
\midrule
\multicolumn{5}{l}{\textit{Proprietary models}} \\[1pt]
\quad Qwen3.5-Plus            & 48.6 & 61.8 & 39.0 & 5.70 \\
\quad GPT-5.4                 & 35.9 & 51.4 & 28.6 & 6.14 \\
\quad Gemini-3.1-Pro-Preview  & 48.5 & 56.4 & 39.9 & 5.59 \\
\quad xAI Grok 4              & 45.5 & 64.8 & 33.4 & 6.18 \\
\quad Claude Opus 4.6         & 21.7 & 48.8 & 15.2 & 9.56 \\
\midrule
\multicolumn{5}{l}{\textit{Open-source models}} \\[1pt]
\quad Qwen3.5-397B-A17B       & 45.2 & 63.9 & 33.3 & 6.15 \\
\quad Kimi-K2.5               & 43.9 & 58.6 & 33.6 & 6.49 \\
\quad LLaVA-Video-7B-Qwen2    & 30.2 & 44.8 & 24.3 & 7.91 \\
\quad Qwen3-VL-8B-Instruct    & 33.4 & 46.8 & 25.9 & 7.09 \\
\quad Qwen3.5-9B              & 36.2 & 45.2 & 29.1 & 6.82 \\
\quad GLM-4.1V-9B-Thinking    & 31.1 & 42.9 & 24.4 & 7.37 \\
\quad GLM-4.6V                & 29.9 & 40.6 & 24.0 & 7.11 \\
\midrule
\multicolumn{5}{l}{\textit{Ours}} \\[1pt]
\quad \model\ & \textbf{63.3} & \textbf{70.0} & \textbf{52.7} & \textbf{4.47} \\
\bottomrule
\end{tabularx}
}
\end{table}

NLG scores measure lexical similarity to reference instructions but do not directly reflect navigational utility.
To bridge this gap, we sample all trajectories from R2R-CE \textit{Val Unseen}, generate instructions with each model under the same prompt, and feed the resulting instructions to a frozen CorrectNav~\cite{yu2026correctnav} follower.
Table~\ref{tab:downstream} reports navigation performance under each instruction source.

\model-generated instructions achieve SR of 63.3 and NE of 4.47 m, on par with human-written instructions (SR 61.6, NE 4.53 m) and closely matching their SPL (52.7 vs. 53.3). We note that a single frozen follower cannot establish superiority over human annotations; these numbers indicate the generated instructions reach a quality level comparable to human references for driving downstream navigation.
Among baselines, the best proprietary model Qwen3.5-Plus reaches only SR 48.6, lagging \model by nearly 15 percentage points.
Open-source models span a wide range (SR 29.9--45.2): the largest ones (Qwen3.5-397B-A17B 45.2, Kimi-K2.5 43.9) are competitive with the proprietary models, while the smaller video VLMs fall to 29.9--36.2. Model scale alone does not determine instruction utility, even Claude Opus 4.6 obtains the lowest SR (21.7) of any source.
These results confirm that NLG metrics and downstream navigation performance are positively correlated, and that \model's spatially grounded instructions translate directly into improved follower behaviour.

\vspace{-2pt}
\subsection{Training Gain from Generated Data}
\vspace{-2pt}

\begin{table}[t]
\centering
\small
\caption{CorrectNav (LLaVA-Video-7B backbone) trained under two augmentation settings on R2R-CE \textit{Val Unseen}.}
\label{tab:datagain}
\setlength{\tabcolsep}{3pt}
\scalebox{0.83}{
\begin{tabularx}{\columnwidth}{c@{\hspace{4pt}}X@{\hspace{6pt}}cccc}
\toprule
\textbf{Set.} & \textbf{Training Data}
  & \textbf{SR}$\uparrow$
  & \textbf{OSR}$\uparrow$
  & \textbf{SPL}$\uparrow$
  & \textbf{NE}$\downarrow$ \\
\midrule
A & R2R-CE + RxR-CE (human)                  & 45.1 & 52.3 & 44.6 & 6.20 \\
B & Setting A + \model-generated             & \textbf{48.4} & \textbf{54.2} & \textbf{46.8} & \textbf{5.85} \\
\bottomrule
\end{tabularx}
}
\end{table}

\begin{table}[t]
\centering
\small
\caption{CorrectNav (LLaVA-Video-7B backbone) trained under two augmentation settings on RxR-CE \textit{Val Unseen}.}
\label{tab:data2}
\setlength{\tabcolsep}{3pt}
\scalebox{0.83}{
\begin{tabularx}{\columnwidth}{c@{\hspace{4pt}}X@{\hspace{6pt}}cccc}
\toprule
\textbf{Set.} & \textbf{Training Data}
  & \textbf{SR}$\uparrow$
  & \textbf{OSR}$\uparrow$
  & \textbf{SPL}$\uparrow$
  & \textbf{NE}$\downarrow$ \\
\midrule
A & R2R-CE + RxR-CE (human)                  & 41.2 & 51.1 & 39.6 & 8.34 \\
B & Setting A + \model-generated             & \textbf{44.4} & \textbf{53.2} & \textbf{41.3} & \textbf{7.74} \\
\bottomrule
\end{tabularx}
}
\end{table}

We further assess whether \model-generated instructions provide greater training benefit.
A CorrectNav~\cite{yu2026correctnav} follower (LLaVA-Video-7B~\cite{zhang2024llavavideo} backbone) is trained from scratch under two data settings and evaluated on R2R-CE \textit{Val Unseen} (Table~\ref{tab:datagain}) and RxR-CE \textit{Val Unseen} (Table~\ref{tab:data2}).
Setting~A uses only the original human-annotated R2R-CE and RxR-CE training data.
Setting~B augments Setting~A with 20K instructions generated by our proposed VTInstructor.

Compared to the human-only baseline (Setting A), Setting B improves SR by +3 percent on both benchmarks and reduces NE by 0.35 m on R2R-CE and 0.60 m on RxR-CE \textit{Val Unseen}, demonstrating that \model-generated data provides meaningful training augmentation for downstream navigation agents.

\vspace{-2pt}
\subsection{Human Evaluation on Real-World Navigation Videos}
\vspace{-2pt}
\label{sec:human_eval}

\begin{table}[t]
\centering
\footnotesize
\caption{Human evaluation on real-world navigation videos (1--5 scale, mean$\pm$std over 3 annotators).}
\label{tab:human}
\setlength{\tabcolsep}{3pt}
\scalebox{0.83}{
\begin{tabularx}{\columnwidth}{X@{\hspace{4pt}}c@{\hspace{5pt}}c@{\hspace{5pt}}c@{\hspace{5pt}}c}
\toprule
\textbf{Method}
  & \textbf{Action}
  & \textbf{Landmark}
  & \textbf{Direction}
  & \textbf{Follow.} \\
\midrule
GPT-5.4                       & 2.65 ± 0.08 & 3.55 ± 0.06 & 2.53 ± 0.10 & 2.29 ± 0.06 \\
Qwen3-VL-8B                   & 2.63 ± 0.12 & 3.65 ± 0.09 & 2.67 ± 0.05 & 2.49 ± 0.08 \\
\model\                              & \textbf{4.38 ± 0.05} & \textbf{4.21 ± 0.05} & \textbf{4.25 ± 0.12} & \textbf{4.35 ± 0.04} \\
\bottomrule
\end{tabularx}
}
\end{table}

To test whether our instructions transfer beyond simulation, we collect \textbf{50 first-person room-tour clips from YouTube} covering offices, corridors, and multi-room apartments. For each clip we estimate per-frame depth and camera pose to recover inter-frame geometry, then apply EDTC and render VTP overlays; the baselines (GPT-5.4 and Qwen3-VL-8B-Instruct) receive the same keyframes. Three annotators rate each instruction on a 1--5 scale along four dimensions: \textbf{Action} (movement and turn sequence), \textbf{Landmark} (object references), \textbf{Direction} (heading
and turn angles), and \textbf{Followability} (whether a na\"ive follower could reproduce the route). Annotators see the keyframes \emph{without} overlays, so scores reflect the instruction alone; we report the mean over annotators. \model leads on all four dimensions (Table~\ref{tab:human}). The gains are largest on Action and Direction, where the VTP overlay supplies movement-sequence and turn-angle cues that general-purpose models lack, while Landmark scores are closer since large VLMs already recognise common indoor
objects.

\section{Limitations and Future Work}
\vspace{-2pt}

The VT-GRPO reward is composed entirely of automatic NLG metrics computed against reference instructions. While this avoids the prohibitive cost of running a navigation follower in the loop, the policy is not directly optimised for navigational success rate. Future work could incorporate sparse follower feedback (e.g., success/failure signals from a lightweight frozen follower on a small trajectory buffer) as an additional reward term, more directly bridging instruction quality and downstream navigation performance.

\begin{acks}
This work was supported by the Beijing Natural Science Foundation (L2608141) and the National Natural Science Foundation of China (62136001).
\end{acks}

\bibliographystyle{ACM-Reference-Format}
\bibliography{reference}

@inproceedings{anderson2018r2r,
  title     = {Vision-and-Language Navigation: Interpreting Visually-Grounded Navigation Instructions in Real Environments},
  author    = {Anderson, Peter and Wu, Qi and Teney, Damien and Bruce, Jake and Johnson, Mark and Sunderhauf, Niko and Reid, Ian and Gould, Stephen and van den Hengel, Anton},
  booktitle = {CVPR},
  year      = {2018}
}

@inproceedings{ku2020rxr,
  title     = {Room-Across-Room: Multilingual Vision-and-Language Navigation with Dense Spatiotemporal Grounding},
  author    = {Ku, Alexander and Anderson, Peter and Patel, Roma and Ie, Eugene and Baldridge, Jason},
  booktitle = {EMNLP},
  year      = {2020}
}

@inproceedings{krantz2020r2rce,
  title     = {Beyond the {Nav-Graph}: Vision-and-Language Navigation in Continuous Environments},
  author    = {Krantz, Jacob and Wijmans, Erik and Majumdar, Arjun and Batra, Dhruv and Lee, Stefan},
  booktitle = {ECCV},
  year      = {2020}
}

@inproceedings{chen2022hamt,
  title     = {History Aware Multimodal Transformer for Vision-and-Language Navigation},
  author    = {Chen, Shizhe and Guhur, Pierre-Louis and Tapaswi, Makarand and Schmid, Cordelia and Laptev, Ivan},
  booktitle = {NeurIPS},
  year      = {2021}
}

@inproceedings{fried2018sas,
  title     = {Speaker-Follower Models for Vision-and-Language Navigation},
  author    = {Fried, Daniel and Hu, Ronghang and Cirik, Volkan and Rohrbach, Anna and Andreas, Jacob and Morency, Louis-Philippe and Berg-Kirkpatrick, Taylor and Saenko, Kate and Klein, Dan and Darrell, Trevor},
  booktitle = {NeurIPS},
  year      = {2018}
}

@inproceedings{tan2019envdrop,
  title     = {Learning to Navigate Unseen Environments: Back Translation with Environmental Dropout},
  author    = {Tan, Hao and Yu, Licheng and Bansal, Mohit},
  booktitle = {NAACL},
  year      = {2019}
}

@misc{yang2026navspacenavigationagentsfollow,
      title={NavSpace: How Navigation Agents Follow Spatial Intelligence Instructions}, 
      author={Haolin Yang and Yuxing Long and Zhuoyuan Yu and Zihan Yang and Minghan Wang and Jiapeng Xu and Yihan Wang and Ziyan Yu and Wenzhe Cai and Lei Kang and Hao Dong},
      year={2026},
      eprint={2510.08173},
      archivePrefix={arXiv},
      primaryClass={cs.RO},
      url={https://arxiv.org/abs/2510.08173}, 
}

@inproceedings{wang2022ccc,
  title     = {Counterfactual Cycle-Consistent Learning for Instruction Following and Generation in Vision-Language Navigation},
  author    = {Wang, Hanqing and Liang, Wei and Shen, Jianbing and Van Gool, Luc and Wang, Wenguan},
  booktitle = {CVPR},
  year      = {2022}
}

@inproceedings{wang2023lana,
  title     = {{LANA}: A Language-Capable Navigator for Instruction Following and Generation},
  author    = {Wang, Xiaohan and Wang, Wenguan and Shao, Jiayi and Yang, Yi},
  booktitle = {CVPR},
  year      = {2023}
}

@inproceedings{gopinathan2024sas,
  author    = {Muraleekrishna Gopinathan and Martin Masek and Jumana Abu-Khalaf and David Suter},
  title     = {Spatially-Aware Speaker for Vision-and-Language Navigation Instruction Generation},
  booktitle = {Proceedings of the 62nd Annual Meeting of the Association for Computational Linguistics (Volume 1: Long Papers)},
  year      = {2024},
  address   = {Bangkok, Thailand},
  publisher = {Association for Computational Linguistics},
  pages     = {13601--13614},
  doi       = {10.18653/v1/2024.acl-long.734},
  url       = {https://aclanthology.org/2024.acl-long.734/}
}

@inproceedings{zeng2023kefa,
  author  = {Haitian Zeng and Xiaohan Wang and Wenguan Wang and Yi Yang},
  title   = {KEFA: A Knowledge Enhanced and Fine-grained Aligned Speaker for Navigation Instruction Generation},
  journal = {arXiv preprint arXiv:2307.13368},
  year    = {2023}
}

@inproceedings{wang2022less,
  author    = {Su Wang and Ceslee Montgomery and Jordi Orbay and Vighnesh Birodkar and Aleksandra Faust and Izzeddin Gur and Natasha Jaques and Austin Waters and Jason Baldridge and Peter Anderson},
  title     = {Less is More: Generating Grounded Navigation Instructions from Landmarks},
  booktitle = {Proceedings of the IEEE/CVF Conference on Computer Vision and Pattern Recognition (CVPR)},
  year      = {2022}
}

@inproceedings{kong2024cinstructor,
  title     = {Controllable Navigation Instruction Generation with Chain of Thought Prompting},
  author    = {Kong, Xianghao and Chen, Jinyu and Wang, Wenguan and Su, Hang and Hu, Xiaolin and Yang, Yi and Liu, Si},
  booktitle = {ECCV},
  year      = {2024}
}

@article{fan2025mapinstructor,
  author    = {Sheng Fan and Rui Liu and Wenguan Wang and Yi Yang},
  title     = {Scene Map-based Prompt Tuning for Navigation Instruction Generation},
  booktitle = {Proceedings of the IEEE/CVF Conference on Computer Vision and Pattern Recognition (CVPR)},
  year      = {2025},
  pages     = {6898--6908}
}

@article{fan2024bevinstructor,
  author    = {Sheng Fan and Rui Liu and Wenguan Wang and Yi Yang},
  title     = {Navigation Instruction Generation with {BEV} Perception and Large Language Models},
  booktitle = {European Conference on Computer Vision (ECCV)},
  year      = {2024}
}

@inproceedings{shtedritski2023whatdoes,
  title     = {What Does {CLIP} Know about a Red Circle? {Visual} Prompt Engineering for {VLMs}},
  author    = {Shtedritski, Aleksandr and Rupprecht, Christian and Vedaldi, Andrea},
  booktitle = {ICCV},
  year      = {2023}
}

@misc{yang2023setofmark,
  title        = {Set-of-Mark Prompting Unleashes Extraordinary Visual Grounding in {GPT-4V}},
  author       = {Yang, Jianwei and Zhang, Hao and Li, Feng and Zou, Xueyan and Li, Chunyuan and Gao, Jianfeng},
  year         = {2023},
  howpublished = {arXiv preprint arXiv:2310.11441}
}

@misc{qwen3vl,
  title     = {Qwen3-VL Technical Report},
  author    = {Qwen Team},
  year      = {2025},
  howpublished = {arXiv preprint}
}

@article{shao2024deepseekmath,
  author  = {Zhihong Shao and Peiyi Wang and Qihao Zhu and Runxin Xu and Junxiao Song and Xiao Bi and Haowei Zhang and Mingchuan Zhang and Y. K. Li and Y. Wu and Daya Guo},
  title   = {DeepSeekMath: Pushing the Limits of Mathematical Reasoning in Open Language Models},
  journal = {arXiv preprint arXiv:2402.03300},
  year    = {2024}
}

@inproceedings{qi2020reverie,
  title     = {{REVERIE}: Remote Embodied Visual Referring Expression in Real Indoor Environments},
  author    = {Qi, Yuankai and Wu, Qi and Anderson, Peter and Wang, Xin and Wang, William Yang and Shen, Chunhua and van den Hengel, Anton},
  booktitle = {CVPR},
  year      = {2020}
}

@article{kamath2022newpath,
  author  = {Aishwarya Kamath and Peter Anderson and Su Wang and Jing Yu Koh and Alexander Ku and Austin Waters and Yinfei Yang and Jason Baldridge and Zarana Parekh},
  title   = {A New Path: Scaling Vision-and-Language Navigation with Synthetic Instructions and Imitation Learning},
  journal = {arXiv preprint arXiv:2210.03112},
  year    = {2022}
}

@inproceedings{li2022envedit,
  title     = {{EnvEdit}: Environment Editing for Vision-and-Language Navigation},
  author    = {Li, Jialu and Tan, Hao and Bansal, Mohit},
  booktitle = {CVPR},
  year      = {2022}
}

@inproceedings{yu2026correctnav,
  author    = {Zhuoyuan Yu and Yuxing Long and Zihan Yang and Chengyan Zeng and Hongwei Fan and Jiyao Zhang and Hao Dong},
  title     = {CorrectNav: Self-Correction Flywheel Empowers Vision-Language-Action Navigation Model},
  booktitle = {Proceedings of the AAAI Conference on Artificial Intelligence},
  year      = {2026},
  volume    = {40},
  number    = {22},
  pages     = {18737--18745}
}

@inproceedings{zhang2024llavavideo,
  title     = {{LLaVA-Video}: Video Instruction Tuning With Synthetic Data},
  author    = {Yuanhan Zhang and Jinming Wu and Wei Li and Bo Li and Zejun Ma and Ziwei Liu and Chunyuan Li},
  booktitle = {NeurIPS},
  year      = {2024}
}

@misc{nasiriany2024pivot,
  title        = {{PIVOT}: Iterative Visual Prompting Elicits Actionable Knowledge for {VLMs}},
  author       = {Nasiriany, Soroush and Xia, Fei and Yu, Wenhao and Xiao, Ted and Liang, Jacky and Dasgupta, Ishita and Xie, Annie and Driess, Danny and Wahid, Ayzaan and Xu, Zhuo and others},
  year         = {2024},
  howpublished = {arXiv preprint arXiv:2402.07872}
}

@inproceedings{zheng2025tracevla,
  title     = {TraceVLA: Visual Trace Prompting Enhances Spatial-Temporal Awareness for Generalist Robotic Policies},
  author    = {Zheng, Ruijie and Liang, Yongyuan and Huang, Shuaiyi and Gao, Jianfeng and Daum{\'e} III, Hal and Kolobov, Andrey and Huang, Furong and Yang, Jianwei},
  booktitle = {The Thirteenth International Conference on Learning Representations (ICLR)},
  year      = {2025}
}

@inproceedings{li2025rovi,
  author    = {Yanbang Li and Ziyang Gong and Haoyang Li and Xiaoqi Huang and Haolan Kang and Guangping Bai and Xianzheng Ma},
  title     = {Robotic Visual Instruction},
  booktitle = {Proceedings of the IEEE/CVF Conference on Computer Vision and Pattern Recognition (CVPR)},
  year      = {2025},
  pages     = {12155--12165}
}

\end{document}